\documentclass{article}

\usepackage{times}
\usepackage{graphicx} 
\usepackage{subfigure} 
\usepackage[table,xcdraw]{xcolor}

\usepackage{algorithmic}
\usepackage{algorithm}
\usepackage{amssymb}
\usepackage{amsmath}
\usepackage{lscape,psfrag,algorithm,algorithmic,color,epsfig,subfigure,multicol,multirow,mathrsfs} %
\usepackage{natbib}
\usepackage{float}
\usepackage{hyperref}

\usepackage[accepted]{whi2016}

\icmltitlerunning{Explaining Classification Models Built on High-Dimensional Sparse Data}

\begin{document}

\twocolumn[
\icmltitle{Explaining Classification Models Built on High-Dimensional Sparse Data}

\icmlauthor{Julie Moeyersoms}{julie.moeyersoms@uantwerp.be}
\icmladdress{University of Antwerp,
	Prinsstraat 13, 2000 Antwerp, Belgium}
\icmlauthor{Brian d'Alessandro}{brian@zocdoc.com}
\icmladdress{Zocdoc, 568 Broadway, New York}
\icmlauthor{Foster Provost}{fprovost@stern.nyu.edu}
\icmladdress{New York University (Stern), 44 West Fourth Street, 8-86
New York}
\icmlauthor{David Martens}{david.martens@uantwerp.be}
\icmladdress{University of Antwerp,
	Prinsstraat 13, 2000 Antwerp, Belgium}
\vskip 0.3in
]

\begin{abstract}
Predictive modeling applications increasingly use data representing people's behavior, opinions, and interactions. Fine-grained behavior data often has different structure from traditional data, being very high-dimensional and sparse.  Models built from these data are quite difficult to interpret, since they contain many thousands or even many millions of features. Listing features with large model coefficients is not sufficient, because the model coefficients do not incorporate information on feature presence, which is key when analysing sparse data. In this paper we introduce two alternatives for explaining predictive models by listing important features.  We evaluate these alternatives in terms of explanation ``bang for the buck,", i.e., how many examples' inferences are explained for a given number of features listed. The bottom line: (i) The proposed alternatives have double the bang-for-the-buck as compared to just listing the high-coefficient features, and (ii) interestingly, although they come from different sources and motivations, the two new alternatives provide strikingly similar rankings of important features. 
\end{abstract}


\section{Introduction}
\label{sec:intro}
Recent studies show that fine-grained behavior data \cite{junque2014} can yield accurate predictive models. What you ``like'' on Facebook for example allows the prediction of personal characteristics remarkably well \cite{Kosinski13}, as well as predicting product interest or credit default behavior~\cite{decnudde15}. Using fine-grained behavior data has been shown to build more accurate models than traditional, structured and engineered data, such as socio-demographic data~\cite{martens2016}. However, accurate predictions are just one important facet in the process of developing and assessing predictive models.  Business stakeholders often need to interpret the model or use it to draw insights. As attention shifts toward explaining how and why models make their predictions, modelers need to balance predictive accuracy and explainability. Prior work suggests that when users do not understand the workings of a classification model, they can be reluctant to use it, even if the model is known to improve decision performance~\cite{Kayande09,EDC}. Furthermore, when pushing to deploy machine learning models, we need to consider that stakeholders often need more than just holdout evaluations to justify chaning their decision-making strategies. The need for explanations encompasses various perspectives, including those of managers, customer-facing employees, and the technical team~\cite{EDC}. 



An important aspect of fine-grained behavior data is its very high dimensionality and sparsity~\cite{junque2014}. Returning to our running example: the Facebook like data can be represented as a matrix where each row (data instance) corresponds to a person, and each column (feature) corresponds to a page on Facebook that one can like. If someone likes a page, the entry is 1, and 0 otherwise. There exists a huge number of possible pages to like, and hence there a huge number of dimensions. On the other hand, a particular user will like only a very small proportion of these pages.  
In ``traditional'' data mining (working with non-behavior data), feature selection and dimensionality reduction techniques are often employed to cope with high dimensionality. However, with fine-grained behavior data it has been shown that feature selection can lead to substantially reduced predictive performance~\cite{junque2013predictive}. Further, dimensionality reduction via singular value decomposition often may not improve predictive performance (and often reduce it)~\cite{Clark2015}, and is dubious for improving explainability in any case.

Linear models are often used for large, sparse behavior data (see references above) as these typically achieve relatively strong predictive performance, while being fast both to train and to deploy. The latter benefit is cited as the prime motivating benefit for their use in large-scale production systems \cite{mcmahan2013ad}. When one tries to explain or interpret such models, the natural tendency is to look at the input features (Facebook pages in our example) with the highest coefficients, as one would do with traditional data. For example, Kosinski et al. (2013) find that the best predictors for high intelligence include Facebook pages \textit{``Thunderstorms''}, \textit{``The Colbert Report''}, \textit{``Science''}, and \textit{``Curly Fries''}. 

In this paper we investigate two alternatives for explaining predictive models from sparse behavior data. The intuition is as follows: the coefficients do not take into account the coverage of the feature (how many users actually like the page).  So, {\bf{to explain the predictions of such classification models, we need to consider both the coefficients and the coverage of the features}}.  We do this by aggregating across instance-based explanations, using two very different approaches. We evaluate these alternatives in terms of explanation ``bang for the buck,", i.e., how many examples' inferences are explained for a given number of features listed. The bottom line: (i) The proposed alternatives have double the bang-for-the-buck as compared to just listing the high-coefficient features, and (ii) interestingly, although they come from different sources and motivations, the two new alternatives provide strikingly similar rankings of important features.



\section{From Instance-Based Explanations to Global Ranking}
We propose alternatives to find the best-explaining features by starting from the instance level. We define two different approaches to obtain such instance-level solutions: one based on the minimum set of features without which the model would not have made the prediction (the "evidence counterfactual" or EC for short), and one based on the Shapley value from cooperative game theory. 
Creating a global model ``explanation'' is a simple procedure. EC or Shapley values are first calculated per feature per instance. Afterwards these scores are summed over all instances and normalized (by the total sum). In this way, the coverage per feature is also taken into account. Features with larger weights will generally get higher values when observed on an instance, but the final explanatory importance of a feature will also depend on how often it is seen.


\subsection{The Evidence Counterfactual}

We draw the first alternative from prior work on explaining document classifications~\citep{EDC}. Here, an ``explanation'' is defined as a minimal set of features (words in the prior work), such that removing these features from the instance  changes the predicted class.  Only when all the words in the explanation are removed does the class change (the set is minimal). In our running Facebook example, if Anna would not have liked \textit{``Data Mining''}, \textit{``the Deer Hunter''} and \textit{``I love reviewing ICML papers''}, then her predicted class would change from highly intelligent to medium intelligent. Hence, for this individual user, these three pages are the explanation why she was classified as highly intelligent. The definition used in the paper is as follows~\citep{EDC}:
\newtheorem{mydef}{DEFINITION}
\begin{mydef}
	Consider a document $D$ consisting of $m_{D}$ unique words $W_{D}$ from the vocabulary of $m$ words: $W_{D}$ = ${w_{i}, i = 1,2,...,m_{D}}$, which is classified by classifier $C_{M}$: $D\rightarrow {1,2,...,k}$ as class $c$. An \textit{explanation for document $D$'s classification} is a set $E$ of words such that removing all words in $E$ from the document leads $C_M$ to produce a different classification. Further, an explanation $E$ is minimal in the sense that removing any subset of E does not yield a change in class. 
	Specifically:  \newline 
	$E$ is an explanation for $C_M(D)$ $\iff$\newline
	1. $E \subseteq W_D$  (the words are in the document),\newline
	2. $C_M(D \backslash  E) \neq c$ (the class changes), and\newline
	3. $\nexists E' \subset E : C_M(D \backslash  E') \neq c$ ($E$ is minimal).\newline
	$D \backslash E$ denotes the result of removing the words in $E$ from document $D$.\newline\label{def:EDC}
\end{mydef}
\vspace{-6mm}
One can interpret the minimal subset as the features that caused the prediction to be made.\footnote{This interpretation is based on the I/O behavior of the classifier.  Examining causality more deeply requires assessing whether it actually makes sense in the domain to assume that the observation can be treated as fixed---so that changing one feature does not change another.} 

Returning to our Facebook example, when a linear model is being used, one could argue simply to list the top $k$ Facebook pages with the highest positive weights that appear in the liking history of a certain user as an explanation for the class. $k$ would then be the minimal number of top pages such that removing these $k$ pages leads to a class change. The evidence counterfactual method (EC) produces minimum-sized combinations for linear models by ranking all pages liked by the user according to the product $\beta_{j}\cdot x_{ij}$ where $\beta_{j}$ is the linear model coefficient and $x_{ij}$ a binary vector that denotes whether or not a page was liked by user $i$. The combinations with the top-ranked pages is a combination of smallest size (the proof can be found in~\citep{EDC}). 

However, it could be interesting to find alternative combinations next to the minimum-size subset. A straightforward approach would be to conduct a complete search through the space of all page combinations, starting with one page, and increasing the number of pages until a subset is found. The algorithm starts by checking whether removing one page from the customer's liking history would cause a change in class label. If so, an irreducible subset is found (in the linear case). If the class does not change based on only one page, the algorithm considers all combinations of size 2, 3 and so on. Note that for a liking history of $m_{A}$ pages, a combination of $k$ pages requires $m_{A}!/(m_{A} - k)!$ evaluations. This complete search scales exponentially with the number of Facebook pages. For data sets with a high dimensionality, this is impracticable if one wants to find multiple combinations. Therefore, a greedy implementation was used~\citep{EDC}.  


\subsection{Approximate Shapley Value}
A second way to find the best explaining features for a single instance can be obtained by using concepts from cooperative game theory, and more specifically the Shapley value. A cooperative game is one in which a set of $N$ players engage in a game that results in some non-negative payoff $v$ for the set. 

Let us first define some core concepts before we present the second method:
\begin{enumerate}
	
	\item $N$ is the complete set of players, with cardinality $\|N\|=n$.
	
	\item $S$ is a subset of players, with $\|S\|=s$ and $S \subset N$
	
	\item $v(S)$ is a value function that represents the total utility (money, points, etc.) the set $S$ generates when playing the game
	
	\item $\Delta_i v(S) = v(S \cup \{i\}) - v(S))$, is the marginal utility of adding player $i$ to a set $S$.
	
\end{enumerate}

The Shapley value (SV) is defined on an individual player $i$, and represents how much of the total value $v(N)$ it should be allocated upon realization of the game. Formally, the SV ($\phi_i$) is the expected marginal utility $E[\Delta_i v(S)]$, of adding player $i$ to a set $S$, where $S$ is the first $s$ players taken from each random permutation of $N$. Or, in other words, the Shapley value($\phi_{i}$) of the game $\langle N,v \rangle$ for player $i$ is the average of its marginal contribution to all possible coalitions. This can be expressed as such~\citep{Shapley}:

\begin{eqnarray}
	\phi_i = E[\Delta_i v(S)] &=& \sum_{S\subset N-\{i\}} P(S) \, \Delta_i v(S) \nonumber \\
	&=& \sum_{S\subset N-\{i\}}\frac{s! (n-s-1)!}{n!} \, \Delta_i v(S) \nonumber
\end{eqnarray}

The method for finding a player's Shapley value depends on the definition of the gain function $(v)$. This function is different depending on the type of game, but in our case we will approach this problem as a weighted voting game. A weighted voting game is a type of game consisting of $N$ players, where each player is defined by a weight $w_i$. The payoff for any weighted voting game played by a subset $S$ is defined as~\citep{gametheory}: 

\[ v(S) = \left\{ 
\begin{array}{l l l}
1 & \quad \text{if $\: \sum_{i \in S} w_i > q$}, & \text{for some specified $q$}\\
\\
0 & \quad \text{otherwise} &
\end{array} \right.
\]

In words, if we think of each player as a voter whose vote is worth $w_i$, a game is won if the total sum of weighted votes is greater than some threshold. From this definition, we can define $\Delta_i v(S)$.

\[ \Delta_i v(S) = \left\{ 
\begin{array}{l l}
1 & \quad \text{if $\: \sum_{j \in S, i \notin S} w_j + w_i \ge q > \sum_{j \in S, i \notin S} w_j$}\\
\\
0 & \quad \text{otherwise}
\end{array} \right.
\]

A player's marginal utility is $1$ if that player's vote swung the total above the threshold ($q$). If the threshold was not met, or the value was already above the threshold, player $i$ adds no marginal value. Bringing this back to our core problem of explaining a model, at an instance level, we can think of the features as playing a weighted voting game, where each feature's weight in the game is the weight learned by a linear model \footnote{Note that this current setup only applies to linear models with binary features.}. Given a set classification threshold, a feature accumulates value if it ``swings'' the classification from negative to positive given $s$ randomly chosen features within the instance summed up before it. 

We refer the interested reader to \citet{moeyersoms2016} for certain necessary practical details (such as negative weights and approximation methods to address scalability).

\begin{figure*}[htb]
		\centering
		\includegraphics[height=6cm, trim={2cm 0.5cm 0.5cm 0.5cm},clip]{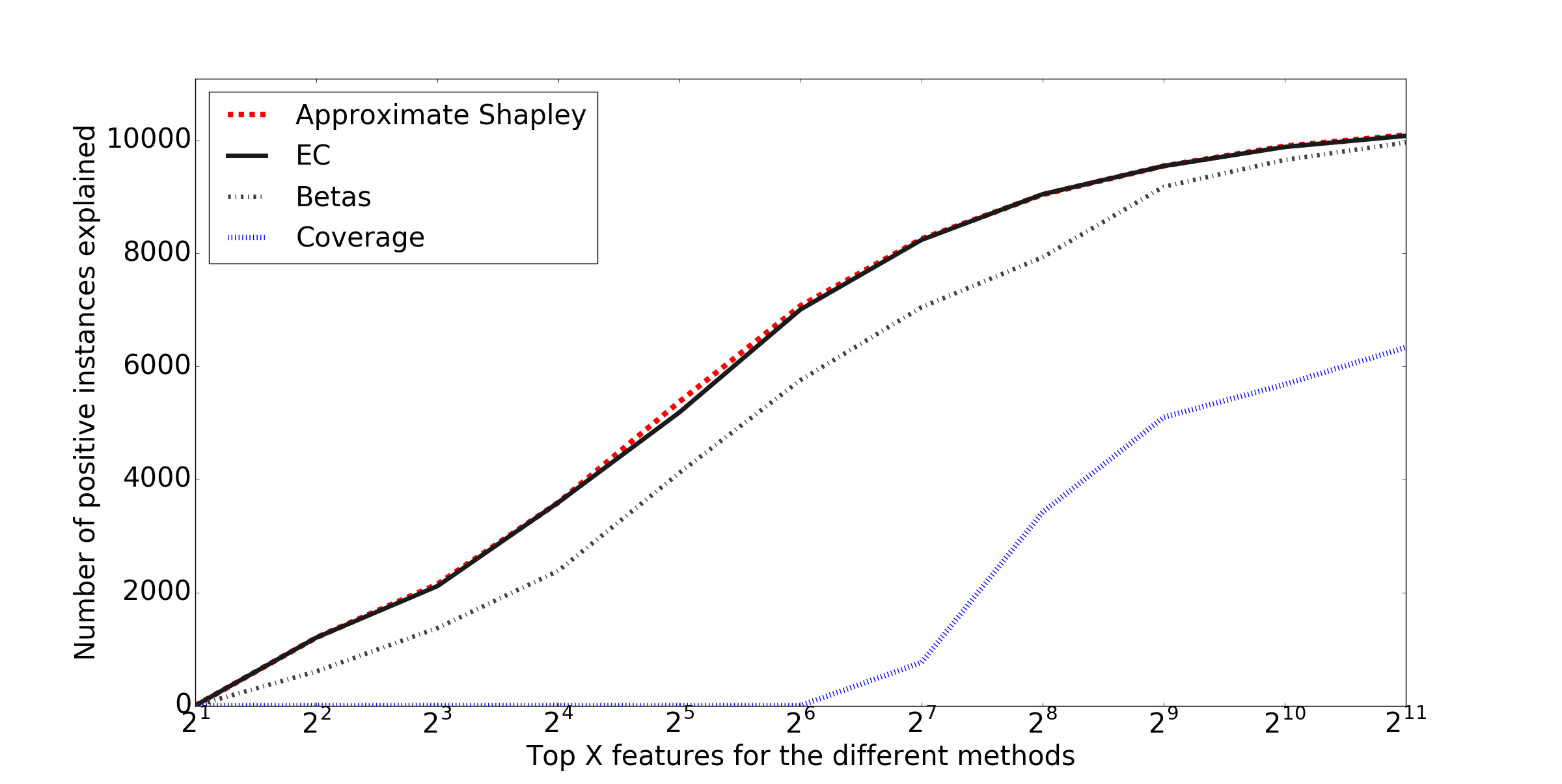}
		\caption{Explanation curves for different ranking approaches.}
		\label{fig:ExplCurve}
\end{figure*}


\section{Empirical Evaluation}
\begin{table*}[]
	\centering
	\caption{Top 10 highest ranked features according to the Evidence Conterfactual (EC), (Approximate) Shapley, $\beta$ and coverage methods.}
	\label{Top20}
	\resizebox{0.99\linewidth}{!}{%
		\begin{tabular}{llll}
			Shapley & EC                & $\beta$  & Coverage \\\hline
			www.bcbg.com  & www.ebay.com                & www.bcbg.com & www.answers.com \\
			www.katespade.com & www.katespade.com       & www.stuartweitzman.com & www.ebay.com \\
			www.ebay.com  & www.bcbg.com                & www.katespade.com & www.huffingtonpost.com \\
			www.stuartweitzman.com & www.stuartweitzman.com   & www.talbots.com & abcnews.go.com \\
			www.restorationhardware.com & www.restorationhardware.com & us.christianlouboutin.com & www.about.com \\
			www.huffingtonpost.com   & www.gilt.com                & www.dior.com & www.forbes.com \\
			www.gilt.com  & www.huffingtonpost.com   & www.restorationhardware.com & www.cnn.com \\
			www.wayfair.com  & www.forever21.com     & www.forever21.com & www.legacy.com \\
			www.talbots.com  & www.colehaan.com      & www.brooksbrothers.com & www.weather.com \\
			www.forever21.com  & www.talbots.com     & www.selfridges.com & www.allrecipes.com \\
		\end{tabular}}
	\end{table*}
	
	For the empirical evaluation, consider predicting product interest based on online browsing data, where a data instance corresponds to an online user, and a feature corresponds to a website. For each user (customer) the feature vector shows the websites
	visited by that user. One typical application of such data is targeted online advertising: who should you target with a certain ad, given the history of all the websites visited by the users. This data is characterized by its high dimensionality and feature sparsity.

	The advertising example that we are using is one from a luxury retail store. The data set consists of several million binary features which respresent URLs visited by each customer. We assume a linear classification model is given, such as logistic regression (see \citet{dalessandro2014scalable}). 
	
	The above EC and Shapley methods provide us ways to rank the features, but how can we empirically evaluate what is to some degree a subjective task? \textbf{We propose ``Explanation curves'', which show how many data instances are explained if one considers only the top ranked features}. The X-axis denotes the number of "top" (top-$k$) features according to that ranking method (in log scale), and the Y-axis shows the number of instances that would be correctly classified as positive (i.e. get a score larger than the threshold) when only those $k$ features are used and all other features are set to zero. 
	
	Figure~\ref{fig:ExplCurve} compares the explanation curves for both methods, as well as for choosing the coefficients with the largest coefficients (Betas; $\beta$), and choosing the terms with the highest coverage.
	One can see that only taking into account the largest coefficients ($\beta$s) of the prediction model explains only half the instances, for almost any point on the explanation curve---a feature may have a very high weight in the prediction model but rarely appears in an instance. Choosing by coverage accounts for this effect, as do both the EC and (Approximate) Shapley value methods. 
	
	Next, Table~\ref{Top20} shows the top 10 highest ranked features according to the different methods. This would be the typical way of explaining such a model (usually with $k$ being larger than 10). As can be seen from these results, ``ebay.com'' is ranked first by the EC method, implying that this is the best explaining feature according to this method. Although this feature has a large coverage, it only appears to be ranked on the 17th place in terms of its $\beta$. Yet, the EC method takes into account both and therefore the ranking will be different. Moreover, when considering the explanatory value of ``ebay.com'', it seems that this value is about 40 times larger according to the EC method (7\%) as compared to just looking at the $\beta$ (proportionally) of the URL (0.17\%). Lastly, and possibly most interestingly of all, Table~\ref{spearmanmethod} shows the correlations between the top 1000 ranked features. Shapley and EC are almost identical in the rankings they provide (as was seen from Table~\ref{Top20} as well). The correlation with $\beta$ however is much smaller.
	
	\begin{table}[H]
		\centering
		\caption{Spearman's rank correlation coefficients between the different methods.}
		\label{spearmanmethod}
		\begin{tabular}{lllll} \hline
			&  Shapley                                  & EC                                      & $\beta$                                              & Coverage                                          \\ \hline
			\multicolumn{1}{l|}{Shapley}  & \multicolumn{1}{l}{\textbackslash}               & \multicolumn{1}{l}{\cellcolor[HTML]{ACACAC}0,97} & \multicolumn{1}{l}{\cellcolor[HTML]{ECECEC}0,60}  & \multicolumn{1}{l}{\cellcolor[HTML]{ECECEC}0,55} \\
			\multicolumn{1}{l|}{EC} & \multicolumn{1}{l}{\cellcolor[HTML]{ACACAC}0,97} & \multicolumn{1}{l}{\textbackslash}               & \multicolumn{1}{l}{\cellcolor[HTML]{ECECEC}0,62} & \multicolumn{1}{l}{0,50}                          \\ 
			\multicolumn{1}{l|}{$\beta$}         & \multicolumn{1}{l}{\cellcolor[HTML]{ECECEC}0,74} & \multicolumn{1}{l}{\cellcolor[HTML]{ECECEC}0,75} & \multicolumn{1}{l}{\textbackslash}               & \multicolumn{1}{l}{0,39}                         \\ 
			\multicolumn{1}{l|}{Coverage}     & \multicolumn{1}{l}{0,16}                         & \multicolumn{1}{l}{0,16}                         & \multicolumn{1}{l}{0,05}                         & \multicolumn{1}{l}{\textbackslash}               
		\end{tabular}
	\end{table}

\section{Conclusion}
We introduced two alternatives for explaining predictive models by listing important features, one drawn from prior work on explaining document classifications, the other derived from the Shapley value used in cooperative game theory.  
We evaluate these alternatives in terms of explanation ``bang for the buck,", i.e., how many examples' inferences are explained for a given number of features listed, as illustrated by ``explanation curves.'' The bottom line conclusions are: (i) Across almost the entire range of the explanation curves, the new alternative explanation methods have double the bang-for-the-buck as compared to just listing the high-coefficient features.  (ii) Very interestingly, although they are derived from quite different sources and motivations, the two new alternatives provide strikingly similar rankings of important features.

\bibliographystyle{icml2016}

\end{document}